\documentclass{bmvc2k}

\title{Prompt-Informed Reinforcement Learning for Visual Coverage Path Planning}

\addauthor{Venkat Margapuri}{vmargapu@villanova.edu}{1}

\addinstitution{
 Villanova University\\
 Department of Computer Science\\
 Villanova, PA
}

\runninghead{Author}{Venkat Margapuri}

\usepackage{amsmath}
\usepackage{amsfonts}

\usepackage{graphicx}
\usepackage{adjustbox}
\usepackage{caption}
\usepackage{wrapfig}

\begin{document}

\maketitle

\begin{abstract}
Visual coverage path planning with unmanned aerial vehicles (UAVs) requires agents to strategically coordinate UAV motion and camera control to maximize coverage, minimize redundancy, and maintain battery efficiency. Traditional reinforcement learning (RL) methods rely on environment-specific reward formulations that lack semantic adaptability. This study proposes Prompt-Informed Reinforcement Learning (PIRL), a novel approach that integrates the zero-shot reasoning ability and in-context learning capability of large language models with curiosity-driven RL. PIRL leverages semantic feedback from an LLM, GPT-3.5, to dynamically shape the reward function of the Proximal Policy Optimization (PPO) RL policy guiding the agent in position and camera adjustments for optimal visual coverage. The PIRL agent is trained using OpenAI Gym and evaluated in various environments. Furthermore, the sim-to-real-like ability and zero-shot generalization of the agent are tested by operating the agent in Webots simulator which introduces realistic physical dynamics. Results show that PIRL outperforms multiple learning-based baselines such as PPO with static rewards, PPO with exploratory weight initialization, imitation learning, and an LLM-only controller. Across different environments, PIRL outperforms the best-performing baseline by achieving up to 14\% higher visual coverage in OpenAI Gym and 27\% higher in Webots, up to 25\% higher battery efficiency, and up to 18\% lower redundancy, depending on the environment. The results highlight the effectiveness of LLM-guided reward shaping in complex spatial exploration tasks and suggest a promising direction for integrating natural language priors into RL for robotics.
\end{abstract}

\section{Introduction}
\label{sec:intro}
\quad In recent times, large language models (LLMs) have advanced significantly in their ability to comprehend and generate human-like text. Notable models such as Generative Pretrained Transformer (GPT) \cite{kirsch2022general}, Bidirectional Encoder Representations from Transformers (BERT) \cite{devlin-etal-2019-bert}, and Text-to-Text Transfer Transformers (T5) \cite{10.5555/3455716.3455856} have exhibited remarkable capabilities in tasks ranging from natural language understanding to sophisticated decision-making process. LLMs are widely adopted in different domains including code generation \cite{wang2023review}, customer chatbots \cite{shareef2024enhancing}, and reinforcement learning (RL) \cite{guo2025deepseek, wang2024reinforcement} systems for their context-aware decision-making ability. Simultaneously, unmanned aerial vehicles (UAVs) have been widely adopted across different sectors such as agriculture \cite{toscano2024unmanned}, delivery services \cite{betti2024uav}, and military \cite{gargalakos2024role}. UAVs are immensely useful for visual coverage path planning (VCPP) tasks where the goal is to cover large swathes of land for tasks such as field surveying, terrain mapping, and crop monitoring. These applications require sophisticated decision-making algorithms to optimize navigation and minimize energy consumption while ensuring comprehensive visual coverage.

The use of RL to train autonomous agents enables them to navigate complex environments by making data-driven decisions based on real-time environmental feedback. However, RL paradigms rely heavily on the reward function, whose curation is often non-trivial as it involves balancing competing objectives such as motion, energy usage, and task relevance. Suboptimal reward design can lead to reward hacking \cite{miao2024inform}, where agents exploit spurious strategies that maximize rewards without accomplishing intended goals such as favoring rapid movement over comprehensive environmental exploration in VCPP tasks. Among RL approaches, Proximal Policy Optimization (PPO) \cite{schulman2017proximal} is prevalent due to its training stability and sample efficiency. However, PPO optimizes against an environment-driven reward formulation which encodes task objectives implicitly but lacks semantic adaptability, potentially limiting the policy's ability to generalize across tasks or environments. Hence, the design of a reward function that accounts for dynamic task constraints remains a bottleneck.

LLMs such as OpenAI GPT-3.5 possess zero-shot reasoning \cite{kojima2022large} and in-context learning abilities \cite{dong2022survey}, generating task-relevant responses from structured prompts without requiring finetuning. While LLMs can produce semantically aligned actions, they lack environmental grounding leading to suboptimal performance. This work proposes Prompt-Informed Reinforcement Learning (PIRL), a hybrid framework that uses LLMs not as policy generators but as reward shapers to dynamically guide RL agents. PIRL combines the adaptability and generalization of LLMs with the optimization strength of RL, enabling agents to learn flexible behaviors aligned with high-level semantic intent. Unlike PPO, PIRL yields task-aware generalization through adaptive reward shaping, and unlike LLM-based planners, PIRL learns from feedback grounded in real-time environmental interaction. By shaping the reward rather than prescribing actions, PIRL enables better alignment with task goals while preserving policy flexibility and improvement without requiring explicit finetuning.

This research contributes to the growing body of work in the field of embodied vision using simulated active agents. By adjusting position and camera parameters, the agent (UAV) trained using PIRL performs VCPP by focusing on maximizing visual coverage while minimizing battery usage and redundant scanning. The key contributions of the work are:
\begin{enumerate}
    \item The use of an LLM, OpenAI GPT-3.5, coupled with PPO RL technique to train a UAV for VCPP in 3D virtual environments using OpenAI Gym and Webots.
    \item Introduction of PIRL, a novel reward shaping paradigm in which context-aware LLMs dynamically guide the reward function using structured zero-shot prompting, mitigating risks such as reward hacking.
    \item Leveraging the zero-shot reasoning ability of LLMs to provide high-level guidance for RL without expert demonstrations enabling scalable policy learning with minimal supervision.
\end{enumerate}


\section{Related Work}
\label{related_work}
\textbf{RL for Visual Coverage Path Planning:} Recent advancements in RL have significantly enhanced frameworks for VCPP by integrating camera aware and curiosity driven strategies. Multiple studies ~\cite{sanchez2021vpp, wang2023coverage, zeng2022deep} have integrated vision sensors for agricultural coverage path planning use cases such as fruit picking and terrain mapping. Another study \cite{carvalho2025deep} introduced a unified framework leveraging value-based RL techniques for zero-shot coverage path planning, emphasizing adaptability across varying map sizes and configurations. In the realm of curiosity driven approaches, ~\cite{yin2023agv, das2024proximal} integrated a curiosity mechanism to enhance the performance of PPO method in a realistic 3D virtual environment. Furthermore, \cite{roghair2021vision} proposed a convergence-based approach and a guidance-based approach using a domain network to explore unexplored areas in a 3D environment while minimizing redundancy. The aforementioned research studies serve as the motivation behind the use of the curiosity driven PPO approach with structured zero-shot prompting to train the UAV for VCPP in 3D environments. \newline
\textbf{Using Language for Reward Shaping:} The use of LLMs to map natural language instructions to rewards has been explored by prior research ~\cite{sharma2022correcting, kwon2023reward, hu2023language, yu2023language}. \cite{huang2022language} used LLMs for grounding high-level tasks expressed in natural language to a chosen set of actionable steps, demonstrating the idea of decomposing high-level tasks to mid-level plans. \cite{kwon2023reward} integrated GPT-3 to train objective-aligned RL agents using few-shot prompting for the Ultimatum game and zero-shot prompting for Matrix games. The work by ~\cite{yu2023language} demonstrated how LLMs can be used to interpret natural language instructions and generate corresponding reward functions using a reward translator. Furthermore, the rewards were converted to code to control robots. This study draws inspiration from the reward translator \cite{yu2023language} in its implementation of Prompt-Adaptive Reward Engine (PARE), described in section \ref{sec:methodology}.\newline
\textbf{Imitation Learning:} Imitation learning helps RL agents learn from expert demonstrations of the task in a supervised learning setting. This technique relies on the availability of large labeled datasets that demonstrate the multitude of scenarios associated with learning a task. Imitation learning has been widely explored in related domains such as autonomous driving ~\cite{teng2022hierarchical, bronstein2022hierarchical}, and path planning ~\cite{chen2023transformer, luo2021self}. This study uses imitation learning as one of the baseline approaches to compare the proposed PIRL approach.


\section{Methodology}
\label{sec:methodology}
The VCPP task is modeled as a Markov Decision Process (MDP) given by the tuple \(\mathcal{M} = (S, A, P, R, \gamma)\), where:
\begin{itemize}
\item \(S \subseteq \mathbb{Z}^3 \times \Theta \times \mathbb{R}_{[0,1]}\) is the hybrid state space, where each state \(s_t \in S\) includes the UAV's 3D position \(\mathbf{p}_t = (x_t, y_t, z_t) \in \mathbb{Z}^3\), camera configuration \(\boldsymbol{\theta}_t = (\phi_t, \psi_t, \zeta_t) \in \Theta\) encoding tilt, pan, and zoom respectively, where \(\Theta := \{(\phi, \psi, \zeta) \in \mathbb{Z} \times \mathbb{Z} \times \mathbb{Q} \mid 0 \leq \phi \leq 90,\ \phi \equiv 0 \pmod{5};\ -90 \leq \psi \leq 90, \psi \equiv 0 \pmod{15}, 0.5 \leq \zeta \leq 2.0, \zeta = 0.1k, \ k \in \mathbb{Z}\} \), and (normalized) battery level \(b_t \in \mathbb{R}_{[0,1]}\), where \(\mathbb{R}_{[0,1]} := \{ b \in \mathbb{R} \mid 0 \leq b \leq 1 \}\).

\item \( A_{\text{move}} = ( x^+, x^-, y^+, y^-, z^+, z^- ) \) and \( A_{\text{cam}} = ( \text{tilt}^+, \text{tilt}^-, \text{pan}^+, \text{pan}^-, \text{zoom}^+, \text{zoom}^- ) \) denote the ordered unit movement and camera control (fixed step sizes of \( 5^\circ \)(tilt), \( 15^\circ \)(pan), and \( 0.1 \)(zoom)) actions, respectively. The discrete action space is defined as the ordered tuple \( A = A_{\text{move}} \mathbin{||} A_{\text{cam}} \), where \( || \) denotes tuple concatenation.

    \item \( P : S \times A \times S \to [0, 1] \) models the probability \( P(s_{t+1} \mid s_t, a_t) \) of transitioning to \( s_{t+1} \) \text{ after executing } \( a_t \) \text{ in } \( s_t \), accounting for deterministic actions and stochastic perturbations like winds.

\end{itemize}

\begin{itemize}

\item \( R : S \times A \rightarrow \mathbb{R} \) is the reward function composed of multiple components designed to encourage exploratory and efficient behavior. The total reward received at time \( t+1 \), resulting from taking action \( a_t \) in state \( s_t \) at time \(t\), is defined as: \newline
\( r_{t+1} = \lambda_c \cdot \Delta C_t + \lambda_r \cdot \mathbb{1}_{\text{redundant}}(s_{t+1}) + \lambda_b \cdot f_{\text{battery}}(b_{t+1}) + \lambda_{\text{cam}} \cdot f_{\text{cam}}(\boldsymbol{\theta}_{t+1}) + \lambda_{\text{cur}} \cdot f_{\text{curiosity}}(s_{t+1}) + \lambda_{\text{collision}} \cdot \mathbb{1}_{\text{collision}}(s_{t+1}) + \lambda_{\text{idle}} \cdot \mathbb{1}_{\text{idle}}(s_{t+1}) + \lambda_{\text{LLM}} \cdot f_{\text{LLM}}(s_{t+1}) \)
where
\begin{itemize}
  \item \( \Delta C_t \): change in visual coverage between timesteps \(t\) and \(t+1\) computed on the ground plane \(z=0\) to prioritize surface-level exploration.
  \item \( \mathbb{1}_{\text{redundant}}(s_{t+1}) \), \( \mathbb{1}_{\text{collision}}(s_{t+1}) \), \( \mathbb{1}_{\text{idle}}(s_{t+1}) \): binary indicators for redundant views, collisions, and idle behavior respectively in \(s_{t+1}\).
  \item \( f_{\text{battery}}(b_{t+1}) \): energy consumption penalty for battery usage observed at \(t+1\).
  \item \( f_{\text{cam}}(\boldsymbol{\theta}_{t+1}) \): incentive to encourage useful variations to pan, tilt, and zoom camera parameters in \(s_{t+1}\).
  \item \( f_{\text{curiosity}}(s_{t+1}) \): reward for the exploration of under-visited areas in \(s_{t+1}\).
  \item \( f_{\text{LLM}}(s_{t+1}) \): auxiliary reward component from PARE that evaluates deviation in position and camera parameters from LLM-based action recommendation at time \(t+1\).
  \item The $\lambda$ coefficients are tunable non-negative scalars \((\mathbb{R}_{\geq 0})\) to balance multiple behavioral objectives.
\end{itemize}

\item \(\gamma \in [0,1]\) is the discount factor that indicates how much future rewards are valued.
\end{itemize}

This work emphasizes visual cell coverage derived from the camera's configuration unlike traditional path planning which rewards cell visitation within the environment. At each step of exploration, the UAV observes a subset of the 3D environment using a forward-facing view cone defined by the camera's configuration. At time \(t\), \(\mathbf{p}_{t} = (x_t, y_t, z_t) \) and \(\boldsymbol{\theta}_t = (\phi_t, \psi_t, \zeta_t) \) describe the UAV's 3D position and camera parameters for (tilt, pan, zoom) respectively. Let \(\text{ViewCone}: \mathbb{Z}^2 \times \Theta \to 2^{\mathbb{Z}^3} \) be the function that maps a given UAV position and camera configuration to a set of observed cells in the 3D environment. At time \(t\), the set of observed cells is defined as \(\mathcal{V}_t = \left\{ (i, j, k) \in \mathbb{Z}^3 \,\middle|\, (i, j, k) \in \text{ViewCone}(\mathbf{p}_t, \boldsymbol{\theta}_t) \right\}\). These observations are used to update the visual coverage map \( \mathcal{C}_{i, j, k}^{(t)} \in \{0, 1\} \) at time \(t+1\) as: \[\mathcal{C}_{i, j, k}^{(t+1)} = 
\begin{cases}
1 & \text{if} (i, j, k) \in \mathcal{V}_t \\
\mathcal{C}_{i, j, k}^t & \text{otherwise}
\end{cases}
\]

\begin{figure}[h]
    \centering
    \includegraphics[width=0.89\linewidth]{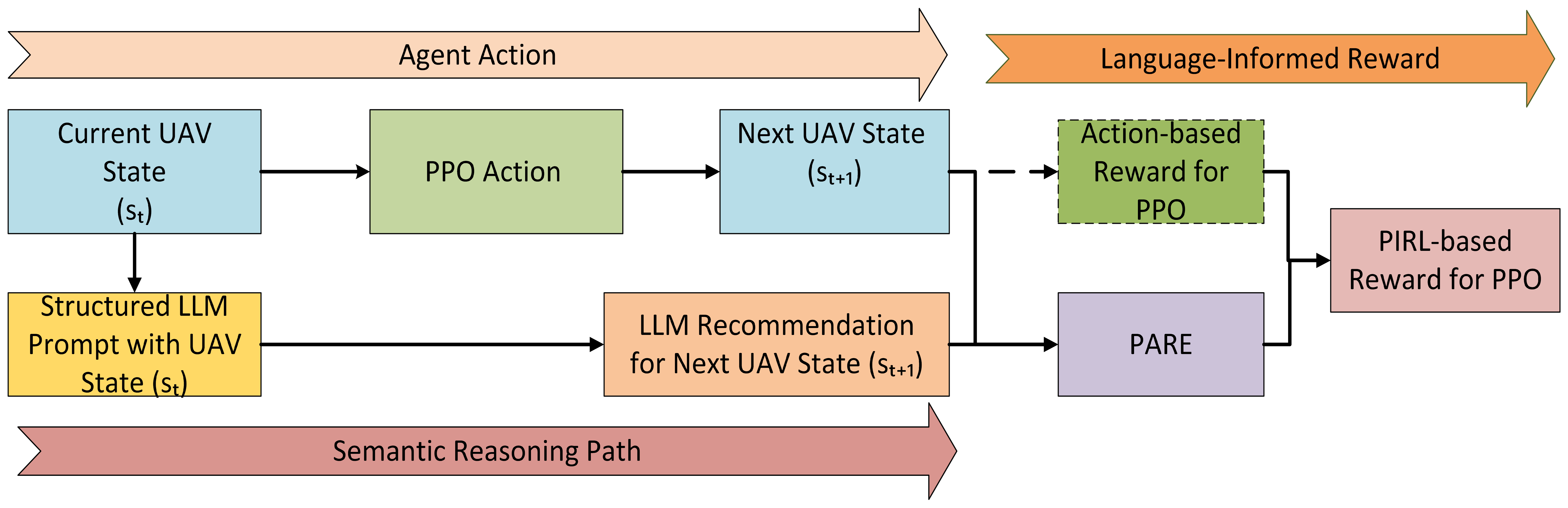}
    \caption{Prompt-Informed Reinforcement Learning for Visual Coverage Path Planning}
    \label{fig:methodology}
\end{figure}

The agent policy \(\pi_\theta(a |s) \) is learned using PPO which maximizes the expected discounted return \(\displaystyle J(\theta) = \mathbb{E}_{\pi_\theta} \left[ \sum_{t=0}^\infty \gamma^t r_t \right]\) where \(r_t\) is the reward at time \(t\) and \(\gamma \in [0, 1]\). PPO ensures that the policy updates are stable using the clipped surrogate objective which clips the policy change at each optimization step. It is given as: 
\(\displaystyle
\begin{aligned}
L^{\text{CLIP}}(\theta) = {} & \mathbb{E}_t \Bigg[ \min \Big( r_t(\theta) \hat{A}_t, \\
& \text{clip}\big(r_t(\theta), 1-\epsilon, 1+\epsilon\big) \hat{A}_t \Big) \Bigg]
\end{aligned}
\) where \( r_t(\theta) = \frac{\pi_\theta(a_t \mid s_t)}{\pi_{\theta_{\text{old}}}(a_t \mid s_t)} \) and \( \hat{A}_t \) is the Generalized Advantage Estimation (GAE). \newline
\null \quad Unlike prior research \cite{yu2023language} where LLM generates task-specific reward functions, this work proposes prompt-informed reinforcement learning (PIRL), a technique that supplements the curiosity-driven action-based reward signal of the PPO RL policy with zero-shot reasoning feedback from an LLM. As illustrated in Figure \ref{fig:methodology}, PIRL dynamically shapes the reward signal at each step of learning based on an LLM's semantic guidance, introducing an adaptive feedback mechanism beyond conventional RL reward formulations. 

A structured text prompt  \(\pi(s_t)\), comprising the concatenation of the following three components is generated using a deterministic parser: 1) task description \(\displaystyle \pi_{\text{task}}\), apprising the LLM of the VCPP task, 2) Environment summary \(\pi_{\text{env}}\), including the UAV's 3D position, camera parameters, battery state, obstacles, and coverage, 3) Request template \(\pi_{\text{query}}\), requesting the LLM for recommendations on the UAV's position and camera parameters. Therefore, \(\displaystyle \pi(s_t) = \pi_{\text{task}} || \pi_{\text{env}} || \pi_{\text{query}}\) where \(||\) denotes string concatenation. Given the structured text prompt \(\pi(s_t)\) at time \(t\), the LLM acts as a mapping function \(\text{LLM}: \mathcal{L} \rightarrow \mathbb{Z}^3 \times \Theta\) generating the recommendation \(\mathbf{a}_{t+1}^{\text{LLM}} = (\mathbf{p}_{t+1}^{\text{LLM}}, \boldsymbol{\theta}_{t+1}^{\text{LLM}})\) where \(\mathbf{p}_{t+1}^{\text{LLM}} = (x_{t+1}^{\text{LLM}}, y_{t+1}^{\text{LLM}}, z_{t+1}^{\text{LLM}}) \in \mathbb{Z}^3\), and \( \boldsymbol{\theta}_{t+1}^{\text{LLM}} = (\phi_{t+1}^{\text{LLM}}, \psi_{t+1}^{\text{LLM}}, \zeta_{t+1}^{\text{LLM}}) \in \Theta\) encode the recommendations for the UAV's 3D position and camera parameters (tilt, pan, zoom) respectively, for time \(t+1\). \newline
\null \quad The proposed PARE module integrates LLM recommendations into the RL policy by computing an auxiliary reward that penalizes the deviation in direction, position, and camera alignment between the agent's actions and the LLM's recommendations. The composite movement alignment reward combines both directional and positional alignment. At time \(t + 1 \), let \(\Delta \mathbf{p}_{t+1} = \mathbf{p}_{t+1} - \mathbf{p}_{t}\) and \(\Delta \mathbf{p}_{t+1}^{LLM} = \mathbf{p}_{t+1}^{LLM} - \mathbf{p}_{t}\) be the agent and LLM-recommended movement vectors respectively. The directional alignment, which quantifies the alignment between the agent's trajectory and LLM's recommended direction (independent of distance), is computed as their cosine similarity defined as: \(\text{DirAlign}(s_{t+1}) = \frac{ \Delta \mathbf{p}_{t+1} \cdot \Delta \mathbf{p}_{t+1}^{\text{LLM}} }{ \| \Delta \mathbf{p}_{t+1} \|_2 \, \| \Delta \mathbf{p}_{t+1}^{\text{LLM}} \|_2 }\) where \(\|\cdot\|_2\) denotes Euclidean (L2) norm. The positional alignment is computed using inverse normalized Euclidean distance, given as: \(\text{PosAlign}(s_{t+1}) = 1 - \frac{ \| \mathbf{p}_{t+1} - \mathbf{p}_{t+1}^{\text{LLM}} \|_2 }{d_{\max}}\), where \(d_{\max}\) is the maximum possible distance in the environment. The composite movement alignment reward is given by: \(f_{\text{move}}(s_{t+1}) = \alpha \cdot \text{DirAlign}(s_{t+1}) + (1 - \alpha) \cdot \text{PosAlign}(s_{t+1})\), where \(\alpha \in [0, 1] \) balances the two alignment components. Likewise at time \(t + 1\), the camera alignment penalty is computed as the deviation between the agent camera parameters \(\boldsymbol{\theta}_{t+1}\) and LLM-recommended camera parameters \(\boldsymbol{\theta}_{t+1}^{\text{LLM}} \). It is defined as: \(\displaystyle f_{\text{cam}}(s_{t+1}) = - \|\boldsymbol{\theta}_{t+1} - \boldsymbol{\theta}_{t+1}^{\text{LLM}} \|_1\) where \(\|\cdot\|_1\) is the Taxicab (L1) norm. The overall reward augmentation by PARE is given as: \(\displaystyle f_\text{LLM}(s_{t+1}) = \lambda_\text{cam} \cdot f_\text{cam}(s_{t+1}) + \lambda_\text{move} \cdot f_\text{move}(s_{t+1}) \text{ where }  \lambda_\text{cam}, \lambda_\text{move} \in \mathbb{R}_{\geq 0} \).\newline
This enables the agent to leverage the LLM as a context-aware, soft-task planner dynamically guiding the reward signal for VCPP.

\section{Experiment}
\label{sec:experiment}

\subsection{Research Questions}
\label{sec:researchquestions}
This study is motivated by the following research questions (RQ): \newline
\null \quad RQ1: Can a pretrained LLM effectively inform and enhance the reward shaping of a PPO-based RL policy in 3D visual coverage environments? \newline
\null \quad RQ2: How does prompt informed reward shaping compare with other learning-based baselines in terms of exploration efficiency, battery utilization, and redundancy minimization? \newline
\null \quad RQ3: Can language-guided reward signals improve generalization across previously unseen 3D environments?

\subsection{Agent Training}
\label{sec:modeltrainingandinference}
The VCPP task is simulated in a custom 3D environment of size 15x15x3 built using the OpenAI Gym framework \cite{brockman2016openai}. The UAV is equipped with a logical pan-tilt-zoom (PTZ) camera whose parameters are trained to maximize visual coverage while minimizing redundant observations and battery consumption. At the start of each episode, spherical obstacles, between two and five, are introduced into the environment dynamically to ensure that UAV learns to avoid obstacles as it explores the environment. Furthermore, at each timestep, stochastic wind disturbances are applied to the environment along the X and Y axes to simulate real-world conditions. The agent is trained using the curiosity-driven PPO approach whose reward function is initialized using exploratory weight random initialization (EWRI). EWRI encourages early exploration by introducing stochasticity into the reward function (RF), drawing reward weights (RW) from a distribution at the beginning of each training episode (shown in Table \ref{tab:ewri_weights}), thereby promoting robustness and generalization. The chosen ranges ensure that the relative priorities among RF coefficients are preserved and positive shaping terms such as coverage, curiosity, and camera movement have a significant impact unless strongly offset by violations like collisions or redundancy. The LLM alignment (\(\lambda_\text{LLM}\)) weight range is deliberately set lower than the upper bound of coverage gain to ensure that maximizing visual coverage remains the most important goal.

\begin{wraptable}{r}{0.62\textwidth}
\captionsetup{type=table}
\captionof{table}{ EWRI Ranges for RF Coefficients}
\vspace{0.4em}
\centering
\small
\begin{tabular}{|l|c|l|}
\hline
\textbf{RW} & \textbf{Range} & \textbf{Description} \\
\hline
$\lambda_c$ & $\mathcal{U}(0.5,\ 1.5)$ & coverage gain \\
$\lambda_r$ & $\mathcal{U}(-1.0,\ -0.2)$ & redundancy penalty \\
$\lambda_b$ & $\mathcal{U}(-0.5,\ -0.1)$ & battery efficiency \\
$\lambda_{\text{cam}}$ & $\mathcal{U}(0.2,\ 0.6)$ & PTZ parameter usage \\
$\lambda_{\text{cur}}$ & $\mathcal{U}(0.2,\ 0.5)$ & curiosity incentive \\
$\lambda_{\text{collision}}$ & $\mathcal{U}(-1.5,\ -0.8)$ & collision penalty \\
$\lambda_{\text{idle}}$ & $\mathcal{U}(-0.8,\ -0.3)$ & idle/inactivity penalty \\
$\lambda_{\text{LLM}}$ & $\mathcal{U}(0.2,\ 0.6)$ & LLM alignment \\
\hline
\end{tabular}
\label{tab:ewri_weights}
\end{wraptable}

At each timestep of agent training, the UAV state which includes 3D UAV position, battery state, coverage, and camera parameters, is curated into a structured zero-shot prompt and passed as input to OpenAI's GPT-3.5 LLM. In return, the LLM recommends the future state values for the UAV position along the Cartesian axes, and camera parameters such as pan, tilt, and zoom. The recommendations made by the LLM are not enforced directly, but incorporated into the reward function using PARE.

\subsection{Evaluation Metrics}
\label{sec:evaluationmetrics}

The trained PPO agent is evaluated using the following core metrics which reflect task success and operational efficiency:

\subsubsection{Visual Coverage Ratio (\( \text{VCR} \))}
VCR is defined as the fraction of unique ground cells (cells at \(z = 0\)) in the 3D environment that are seen by the UAV's camera view cone as the camera configuration is altered during an episode. It is given by \(\text{VCR} = \frac{|\mathcal{C}_{\text{covered}}|}{|\mathcal{C}_{\text{total}}|}\) where \( \mathcal{C}_{\text{covered}} \subseteq \mathcal{C}_{\text{total}} \); \( \mathcal{C}_{\text{covered}}\) and \(\mathcal{C}_{\text{total}} \) are the set of uniquely observed ground cells and total ground cells in the environment respectively, for a given episode. The higher the VCR, the better the UAV's coverage of the environment. 

\subsubsection{Battery Efficiency (\( \text{BE} \))}
BE is defined as the VCR per unit battery consumed, and is given by \(\text{BE} = \frac{\text{VCR}}{2 - \frac{b_t}{b_i}} \), where \(b_i\) and \(b_t\) represent the initial and final battery level for the episode. The higher the BE, the better the UAV's battery usage.
\subsubsection{Redundant Views per Covered Cell (\( \text{RVC} \))}
RVC measures spatial redundancy in ground-level observations, and is computed as the average number of redundant views per uniquely covered ground cell. It is given by \(\text{RVC} = \frac{|\mathcal{C}_{\text{redundant}}|}{|\mathcal{C}_{\text{covered}}|} \), where \( \mathcal{C}_{\text{redundant}} \) is the set of ground cells observed more than once during the episode. The lower the \(\text{RVC}\), the better the efficiency in coverage with minimal redundancy.

\subsection{Results}
\label{sec:results}

\begin{wrapfigure}{r}{0.4\textwidth}
  \centering
  \includegraphics[width=0.30\textwidth]{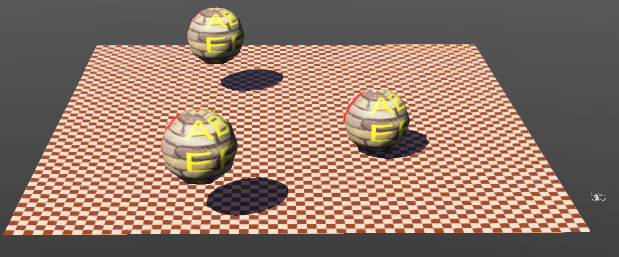}
  \vspace{0.5em} 
  \caption{Environment with Three Obstacles and UAV in Webots}
  \label{fig:Webots Environment}
\end{wrapfigure}

\quad \null The trained PPO agent is inferred in two different simulation environments, OpenAI Gym, and Webots. While OpenAI Gym offers a grid-based, logic-level simulation, Webots allows for physics-based, sensor-level robotic simulation. Therefore, evaluating the model in Webots allows for sim-to-real-like evaluation by determining whether the learned policy can tolerate an imperfect and realistic 3D environment. The agent is evaluated zero-shot in test environments of varying sizes including 30x30x3, 45x45x3, and 60x60x3 with the number of steps and battery strength scaled using the scaling factor \( \displaystyle
\alpha = \frac{x_{\text{test}} \cdot y_{\text{test}} \cdot z_{\text{test}}}{x_{\text{train}} \cdot y_{\text{train}} \cdot z_{\text{train}}}
\). The OpenAI Gym environment is emulated in Webots (Figure \ref{fig:Webots Environment}) by creating a logical mirror of the UAV simulation state. The trained agent is simulated on a DJI Mavic 2 Pro fitted with a controllable gimbal-mounted camera in the Webots simulator. Although Webots does not interpret image streams natively for RL, it allows position-based coverage approximation consistent with the OpenAI Gym implementation. Zero-shot inference is performed in the Webots simulator with the agent not undergoing any training in a Webots environment. The evaluation metrics described in section \ref{sec:evaluationmetrics} are logged across 100 independent episodes for each environment wherein obstacles and UAV start position are varied for each episode, essentially inferring on 100 varied environments of the same size to determine the performance of the trained agent. To ensure consistency, the obstacle locations and start positions in different environments are preserved as the proposed PIRL agent is compared with baseline agents. \newline
\null \quad The inference results of the proposed PIRL approach are compared with multiple baseline approaches. A curiosity-driven PPO baseline with static reward function and no LLM guidance (PPO-SR), and a PPO policy initialized with EWRI without LLM guidance (PPO-EWRI), LLM-only without PPO (LLM-only), and imitation learning (IL) using TabNet \cite{arik2021tabnet} are used to compare with the proposed approach. The mean of each of the evaluation metrics from running inference on 100 episodes in OpenAI Gym and Webots is as shown in Tables \ref{tab:inference_comparison_openaigym} and \ref{tab:inference_comparison_webots} respectively.

\noindent
\begin{minipage}[t]{0.49\textwidth}
\centering
\small
\captionof{table}{Mean of Results in OpenAI Gym}
\label{tab:inference_comparison_openaigym}
\vspace{0.5em}
\begin{adjustbox}{width=\linewidth}
\begin{tabular}{|l|ccc|ccc|ccc|}
\hline
\textbf{Method} & \multicolumn{3}{c|}{\textbf{30x30x3}} & \multicolumn{3}{c|}{\textbf{45x45x3}} & \multicolumn{3}{c|}{\textbf{60x60x3}} \\
\cline{2-10}
& \textbf{VCR} & \textbf{BE} & \textbf{RVC} & \textbf{VCR} & \textbf{BE} & \textbf{RVC} & \textbf{VCR} & \textbf{BE} & \textbf{RVC} \\
\hline
PPO-SR         & 0.64 & 0.55 & 0.22 & 0.61 & 0.56 & 0.24 & 0.63 & \textbf{0.48} & 0.26 \\
PPO-EWRI       & 0.61 & 0.59 & 0.24 & 0.59 & \textbf{0.62} & 0.25 & 0.62 & 0.36 & 0.27 \\
LLM-only       & 0.56 & 0.42 & 0.28 & 0.53 & 0.40 & 0.31 & 0.55 & 0.39 & 0.27 \\
IL             & 0.48 & 0.21 & 0.40 & 0.44 & 0.15 & 0.47 & 0.41 & 0.11 & 0.49 \\
PIRL (Ours) & \textbf{0.73} & \textbf{0.74} & \textbf{0.18} & \textbf{0.71} & 0.58 & \textbf{0.21} & \textbf{0.70} & 0.23 & \textbf{0.24} \\
\hline
\end{tabular}
\end{adjustbox}
\end{minipage}
\hfill
\begin{minipage}[t]{0.49\textwidth}
\centering
\small
\captionof{table}{Mean of Results in Webots}
\label{tab:inference_comparison_webots}
\vspace{0.5em}
\begin{adjustbox}{width=\linewidth}
\begin{tabular}{|l|ccc|ccc|ccc|}
\hline
\textbf{Method} & \multicolumn{3}{c|}{\textbf{30x30x3}} & \multicolumn{3}{c|}{\textbf{45x45x3}} & \multicolumn{3}{c|}{\textbf{60x60x3}} \\
\cline{2-10}
& \textbf{VCR} & \textbf{BE} & \textbf{RVC} & \textbf{VCR} & \textbf{BE} & \textbf{RVC} & \textbf{VCR} & \textbf{BE} & \textbf{RVC} \\
\hline
PPO-SR          & 0.61 & \textbf{0.47} & 0.21 & 0.62 & 0.52 & 0.26 & 0.65 & 0.30 & 0.22 \\
PPO-EWRI        & 0.63 & 0.44 & 0.19 & 0.65 & 0.57 & 0.24 & 0.60 & 0.32 & 0.24 \\
LLM-only        & 0.59 & 0.40 & 0.26 & 0.57 & 0.43 & 0.24 & 0.58 & \textbf{0.36} & 0.23 \\
IL              & 0.36 & 0.29 & 0.39 & 0.29 & 0.22 & 0.41 & 0.37 & 0.30 & 0.44 \\
PIRL (Ours) & \textbf{0.70} & 0.42 & \textbf{0.14} & \textbf{0.79} & \textbf{0.60} & \textbf{0.22} & \textbf{0.71} & 0.35 & \textbf{0.17} \\
\hline
\end{tabular}
\end{adjustbox}
\end{minipage}

\subsection{Discussion}
\label{sec:discussion}
\quad RQ1: Can a pretrained LLM effectively inform and enhance the reward shaping of a PPO-based RL policy in 3D visual coverage environments? \newline
\null \quad The results from OpenAI Gym and Webots environments demonstrate that the LLM-guided PIRL significantly enhances the visual coverage capabilities of the PPO agent. In OpenAI Gym, on the 30x30x3 environment, PIRL achieves a VCR of 0.73 compared to 0.64 (PPO-SR), 0.61 (PPO-EWRI), 0.56 (LLM-only), and 0.48 (IL), exhibiting relative improvements ranging from 14\% (over PPO-SR) to 52\% (over IL). Similarly, PIRL records VCR of 0.71 and 0.70 in larger environments of sizes 45x45x3 and 60x60x3 respectively, outperforming the best baseline (PPO-SR) by 16\% and 11\% respectively. Furthermore, a similar trend is observed in the Webots simulator which introduces realistic physical dynamics and environmental perturbations such as wind.  In the 30x30x3 environment, PIRL scores a VCR of 0.70 surpassing PPO-EWRI (0.63), PPO-SR (0.61), LLM-only (0.59), and IL (0.36). It translates to performance gains ranging from 11\% (over PPO-EWRI) to 94\% (over IL). Likewise, PIRL achieves a VCR of 0.79 and 0.71 on 45x45x3 and 60x60x3 environments respectively, maintaining margins of 27\% and 9\% over the best baselines of PPO-SR and PPO-EWRI respectively. The results affirm the feasibility of LLM-informed reward shaping to improve the visual coverage ability of a PPO-based RL policy in 3D environments. \newline
\null \quad RQ2: How does prompt informed reward shaping compare with other learning-based baselines in terms of exploration efficiency, battery utilization, and redundancy minimization? \newline
\null \quad The proposed PIRL approach improves the exploration efficiency of the PPO agent by guiding behavior that minimizes redundancy and promotes diverse viewpoint adjustments. This is evidenced by the best RVC values achieved by PIRL across both OpenAI Gym and Webots simulators. In OpenAI Gym, PIRL achieves RVC of 0.18, 0.21, and 0.24 in the 30x30x3, 45x45x3, and 60x60x3 environments respectively.  They are lower (better) than those achieved by the best baseline, PPO-SR, which achieves RVC scores of 0.22, 0.24, and 0.26 on the same environments translating to relative improvements of 18\%, 4\%, and 8\% respectively in redundancy. Similarly, in Webots, PIRL achieves the best RVC scores of 0.14, 0.22, and 0.17 in the 30x30x3, 45x45x3, and 60x60x3 environments, outperforming all the baseline approaches. \newline
\null \quad BE results offer more nuanced trends compared to RVC. In OpenAI Gym, PIRL attains a BE of 0.74 in the 30x30x3 environment, 25\% higher than the nearest baseline (PPO-EWRI - 0.59) but exhibits lower BE in larger environments. Specifically, PIRL’s BE drops to 0.58 in the 45x45x3 environment (lower than PPO-EWRI’s 0.62) and 0.23 in the 60x60x3 environment, trailing PPO-SR (0.48), PPO-EWRI (0.36), and LLM-only (0.39). Despite this drop, PIRL continues to achieve the highest VCR which indicates that PIRL favors maximizing energy utilization for diverse observations rather than minimizing consumption, a learned preference shaped by dynamic reward signals. In Webots, PIRL maintains a competitive or superior BE profile, achieving 0.60 on the 45x45x3 grid, the highest among all approaches. Though it is slightly outperformed in the \text{30x30x3} and \text{60x60x3} environment, PIRL’s superior RVC and VCR values across all grid sizes compensate for the minor BE trade-off. Therefore, PIRL encourages the PPO agent to explore meaningfully capturing more unique observations with a reasonable trade-off between coverage and energy expenditure.\newline
\null \quad RQ3: Can language-guided reward signals improve generalization across previously unseen 3D environments? \newline
\null \quad As part of the experiment, all the agents are initially trained in a 15x15x3 environment using OpenAI Gym. The agents are inferred in environments of sizes 30x30x3, 45x45x3, and 60x60x3 in Webots to test the zero-shot ability of the agents in a realistic physics-based simulator. The robustness of the PIRL agent in Webots, despite being trained exclusively in OpenAI Gym, demonstrates its strong generalization capacity to zero-shot, sim-to-real-like environments. PIRL achieves high VCR and low RVC values across all three grid sizes in Webots without any additional fine-tuning. For instance, PIRL improves VCR over IL by 94\% on the 30×30×3 grid, and maintains strong lead margins of 27\% and 9\% on the 45×45×3 and 60×60×3 grids, respectively, over the best baselines. Furthermore, PIRL achieves the lowest RVC scores in all Webots settings, and while its BE is not the highest in all environments, it reaches a peak of 0.60 BE on the 45×45×3 grid, outperforming all baselines in that configuration. These results validate that the language-guided feedback used for reward modulation in PIRL allows the agent to learn generalizable exploration behaviors that remain effective even in high-fidelity, unstructured environments with physical disturbances.

\subsection{Limitations}
Although the proposed PIRL approach demonstrates robust performance across diverse environments, the approach comes with a few limitations. They are: 
\begin{itemize}
    \item PIRL relies on access to a pretrained LLM with zero-shot reasoning and in-context learning capabilities to generate semantic guidance during training. This dependency may introduce computational overhead and latency, which could be limiting for resource-constrained settings or cost-sensitive applications.
    \item The current formulation models the state space discretely for both position and camera parameters, which simplifies the learning problem but may limit the granularity of control and realism. Extending PIRL to continuous or hybrid state-action spaces would enable more precise control and facilitate real-world deployment. 
    \item While sim-to-real-like evaluation using Webots introduces physical realism, real-world deployment would require further considerations such as sensor noise, localization error, and computational constraints onboard the UAV platform.
\end{itemize}

\section{Conclusion and Future Work}
This study presents PIRL that leverages LLMs for dynamic reward shaping of a PPO policy aimed at visual coverage of 3D environments. Through robust experimental setup and extensive evaluation in OpenAI Gym and Webots, this study demonstrates the superior performance of PIRL for VCPP by comparing with multiple learning-based baseline approaches. While the proposed approach highlights the potential of LLM-guided reward shaping, future work will focus on enhancing PIRL by integrating visual signal fusion where real-time camera streams are semantically interpreted to adaptively influence reward dynamics. Furthermore, incorporating LLM self-evaluation mechanisms such as confidence scores can help modulate the influence of LLM guidance during training in different environments. These research directions will be actively pursued to increase the robustness, and real-world applicability of PIRL-based systems.


\bibliography{bmvc_review}

\end{document}


\appendix
\section{Appendix}
\subsection{State Space Design Rationale}
\label{sec:appendix-state-space-design}
The ranges for tilt, pan, and zoom reflect a balance between computational expressiveness and real-world tilt, pan, and zoom parameters in UAVs. The rationale for each of them is as follows:
\begin{itemize}
    \item The tilt range from $0^{\circ}$ to $90^{\circ}$ allows the camera to shift from forward-facing to downward views, a critical trait for navigation and visual coverage.
    \item The pan ranging from $-90^{\circ}$ to $90^{\circ}$ facilitates symmetric lateral scanning while preventing $360^{\circ}$ rotations. Such a rotation is not only unnecessary in structured tasks like VCPP, but could also lead to state space explosion, complicating control logic.
    \item The zoom range of 0.5x to 2x enables the UAV to alternate between wide-area observation and localized inspection, with a bounded and discrete set of zoom levels that facilitate policy generalization. 

\end{itemize}

While the parameters chosen for the tilt, pan, and zoom might not match the precise physical bounds of commercial UAV cameras, they are sufficient to demonstrate that PIRL can leverage LLM-based semantic feedback to guide low-level control decisions made by an RL policy. Furthermore, binding the state space to physical bounds drives faster policy convergence in simulation settings. The proposed approach is agnostic to the exact parameterization of the state space and could be adapted to match real-world hardware specifications. 

\subsection{PARE Design Rationale}
\label{sec:appendix-PARE-Design-Rationale}

\subsubsection{Design Rationale for LLM-Based Reward Components}
\label{appendix-Design Rationale for LLM-Based Reward Components}
In PARE, the camera alignment is treated as an unrewarded hard constraint whereas the movement alignment is treated as a rewarded soft constraint. This design choice is rooted in the semantics of the VCPP task. The rationale is explained as follows: \newline
\textbf{Camera Alignment as a Hard Constraint:} The camera configuration of the UAV directly determines the shape and direction of the view cone, which in turn affects the ground level visual coverage. Deviations from LLM recommended camera parameters may lead to substantial reductions in observable area as coverage opportunities are often transient due to the UAV being in continuous motion. Therefore, deviations from LLM recommendations are penalized but positive rewards are not provided for compliance. Instead, the agent is expected to follow the recommendations made by the LLM as constraints essential to successful perception.  \newline
\textbf{Movement Alignment as a Soft Constraint:} Movement recommendations made by the LLM reflect high-level positional guidance but are flexible due to real-time environmental dynamics such as obstacles, wind, and occlusion. Since position changes are more tolerant to perturbations, the movement alignment is modeled as a soft constraint rewarding the agent for complying with LLM guidance but not strictly penalizing minor deviations, thereby allowing the RL policy to balance LLM recommendations with its own learned behavior in a dynamic environment.

\subsubsection{Decoupling Direction and Position for Alignment for Composite Movement Alignment}
Direction and position encode orthogonal aspects of alignment. Directional alignment evaluates whether the agent moves toward the LLM's recommended direction, regardless of how far it travels. This is useful to encourage trajectory-level semantic consistency. In contrast, positional alignment evaluates whether the agent reaches the target location recommended by the LLM. This is useful to encourage spatial goal realization of the agent. 
The complementary terms of directional and positional alignment are incorporated together as the composite movement alignment reward. A tunable coefficient \(\alpha \in [0, 1]\) is used to weight their contributions. The use of a convex combination ensures that the resulting reward remains within a bounded and interpretable range. The constraint \(\alpha \in [0, 1]\) allows the agent to interpolate smoothly between emphasizing directional agreement \(\alpha \to 1\) and positional proximity \(\alpha \to 0\), thereby justifying the choice.

\subsection{Why GPT-3.5?}
OpenAI GPT-3.5 is chosen for this study despite newer models such as OpenAI GPT-4 being available for the following reasons:
\begin{itemize}
    \item \textbf{Cost and Token Efficiency:} Compared to GPT-4, GPT-3.5 provides a more economical inference cost and faster response latency which are important when processing a large number of prompts over multiple training episodes.
    \item \textbf{Sufficient Reasoning Ability:} The task of VCPP requires semantic grounding and basic spatial reasoning rather than deep abstract problem solving. GPT-3.5 is perfectly capable of doing so, especially when coupled with structured zero-shot prompting.
    \item \textbf{Empirical Validation:} Pilot experiments using GPT-3.5 yielded meaningful and context-aligned recommendations justifying its use throughout the training process of the experiment.
\end{itemize}

\subsection{Structured Zero-Shot Prompt}
\label{appendix:prompt}
The structured zero-shot prompt that provides the UAV's position and camera parameters of pan, tilt, and zoom consists of three concatenated components:

\begin{itemize}
    \item \textbf{Task Description \((\pi_{\text{task}}\)):} The task description instructs the LLM of the VCPP task which is to maximize ground coverage while avoiding obstacles, minimizing battery usage and redundant scanning of the environment.
    \item \textbf{Environment Summary \((\pi_{\text{env}}\)):} The environment summary encodes the UAV's current state:
    \begin{itemize}
        \item \texttt{UAV Position:\([x, y, z]\)}
        \item \texttt{Camera Parameters:\([\text{field of view}, \text{ resolution}, \text{ tilt}, \text{ pan}, 
        \\ \text{ zoom}]\) }
        \item \texttt{Battery Level}
        \item \texttt{Coverage}
        \item \texttt{View Cone Information}
        \item \texttt{Obstacle Information}
    \end{itemize}
    \item \textbf{Request Template, (\(\pi_{\text{query}}\)):} The request template specifies the rules that the LLM needs to adhere as it generates the recommendations. Rules which include the parameters for which the LLM needs to generate recommendations, range of the parameters, and output format for recommendations, enable the LLM to generate template adhering response that are feasible for parsing by PARE.
\end{itemize}

The complete prompt is curated as: \( 
\pi(s_t) = \pi_{\text{task}} \, || \, \pi_{\text{env}} \, || \, \pi_{\text{query}}
\) where \(||\) denotes string concatenation. \newline

\textbf{Sample Structured Zero-Shot Prompt}
\begin{quote}
\ttfamily
[Task Description]\\
You are a drone controller fitted with a gimbal-mounted camera tasked with maximizing visual ground coverage within a 3D environment. Suggest movements in the 3D space (X, Y, Z) and camera adjustments (pan, tilt, zoom) that enhance visual ground coverage while minimizing battery consumption, avoiding obstacles, and reducing redundant observations.\\

[Environment Summary] \\
Drone state: \{'drone\_position': [14, 0, 2], 'camera\_params': [90.0, 1.0, 60.0, 90.0, 1], 'battery': 0.24, 'coverage': 0.7837\}. The drone is currently operating in a 15$\times$15$\times$3 grid environment and has visually covered a portion of the ground-level cells using its camera. The camera params in the drone state are listed in order as: field of view (degrees), resolution, tilt (degrees), pan (degrees), and zoom. The camera emits a downward-facing square view cone projected onto the ground level (z=0), centered on the drone’s current (x, y) location. The size of the view cone is determined by the zoom level and tilt angle. If tilt < 80, the cone has a half-width of approximately \texttt{2$\times$zoom} units in both x and y directions; otherwise, it is \texttt{1$\times$zoom}. There are two spherical obstacles in the \\environment:
1. Location: [5,5,1] with radius: 1.5 units
2. Location: [10, 10, 2] with radius: 2 \\

[Request Template] \\
1. You can adjust tilt, pan, zoom, and movement along the X, Y, and Z axes. \\
2. The battery ranges from 0 to 1 where 1 and 0 are the maximum and minimum battery levels respectively. The drone will be unable to move once battery level reaches 0. \\
3. The values should be in the range: pan: [-90, 90] degrees, tilt: [0, 90] degrees, zoom: [0.5, 2.0], X:[-1, 1], Y:[-1, 1], Z:[-1, 1].
\end{quote}

\textbf{Sample LLM Response:}
\begin{quote}
\ttfamily
pan: 75, tilt: 45, zoom: 1, X: -1, Y: 0, Z: 0
\end{quote}

Please note that the request template only requests the LLM to generate values within the allowable limits for 3D position and camera parameters, but does not expect the recommendations to be discretized values or atomic actions. This is because PIRL does not use the recommended actions to drive the agent but considers the recommendations to shape the reward using PARE. Constraining the LLM would lead to the loss of semantic richness and directionality in the feedback signal. 

\subsection{Baseline Approaches and Results}
\subsubsection{TabNet-based Imitation Learning} 
\label{sec:appendix-bl-training}
TabNet is a deep learning model designed specifically for tabular data, combining sequential attention and feature selection with interpretability. Unlike standard multilayer perceptrons or transformer-based encoders, TabNet dynamically focuses on the most relevant subset of input features at each decision step through a sparse attention mechanism. This makes it particularly well-suited for the VCPP task, which involves structured and low-dimensional state representations. \newline
The imitation learning task is modeled as a multiclass classification problem over 12 discrete atomic actions (six movement and six camera control operations). The training data is collected by logging the state and action at each step from PPO-EWRI rollouts on a \(15 \times 15 \times 3\) environment in OpenAI Gym. Each row in the dataset encodes the full environment state and the corresponding action label in natural language. These action labels are deterministically mapped to discrete class indices, and the input features are standardized prior to training. The model is trained using cross-entropy loss. The trained TabNet model is evaluated in both OpenAI Gym and Webots simulators, and the results are reported in Table~1 and Table~2. This model provides a supervised learning baseline trained on data from a non-LLM expert policy.

\subsubsection{LLM-only Learning}
The LLM-only learning technique is a zero-shot approach where the LLM, GPT-3.5, directly provides the recommendation for action at each decision point, without RL or policy optimization in the loop. In this approach, the LLM acts as a high-level decision making engine by producing semantically informed action recommendations in response to a structured text prompt that encodes the current environment state. The construction of the structured prompt is described in section \ref{appendix:prompt} of the Appendix. The natural language responses of the LLM are parsed and implemented as-is to drive the agent. The LLM is neither finetuned nor provided with examples in the structured prompt. Therefore, this baseline approach provides an estimate of how well semantic priors and language-based reasoning perform in solving the task of VCPP. The agent is evaluated in OpenAI Gym and Webots simulators, and the results are presented in Table 1 and Table 2. 

\subsubsection{Results}
\begin{figure}[H]
    \centering
    \includegraphics[width=0.89\linewidth]{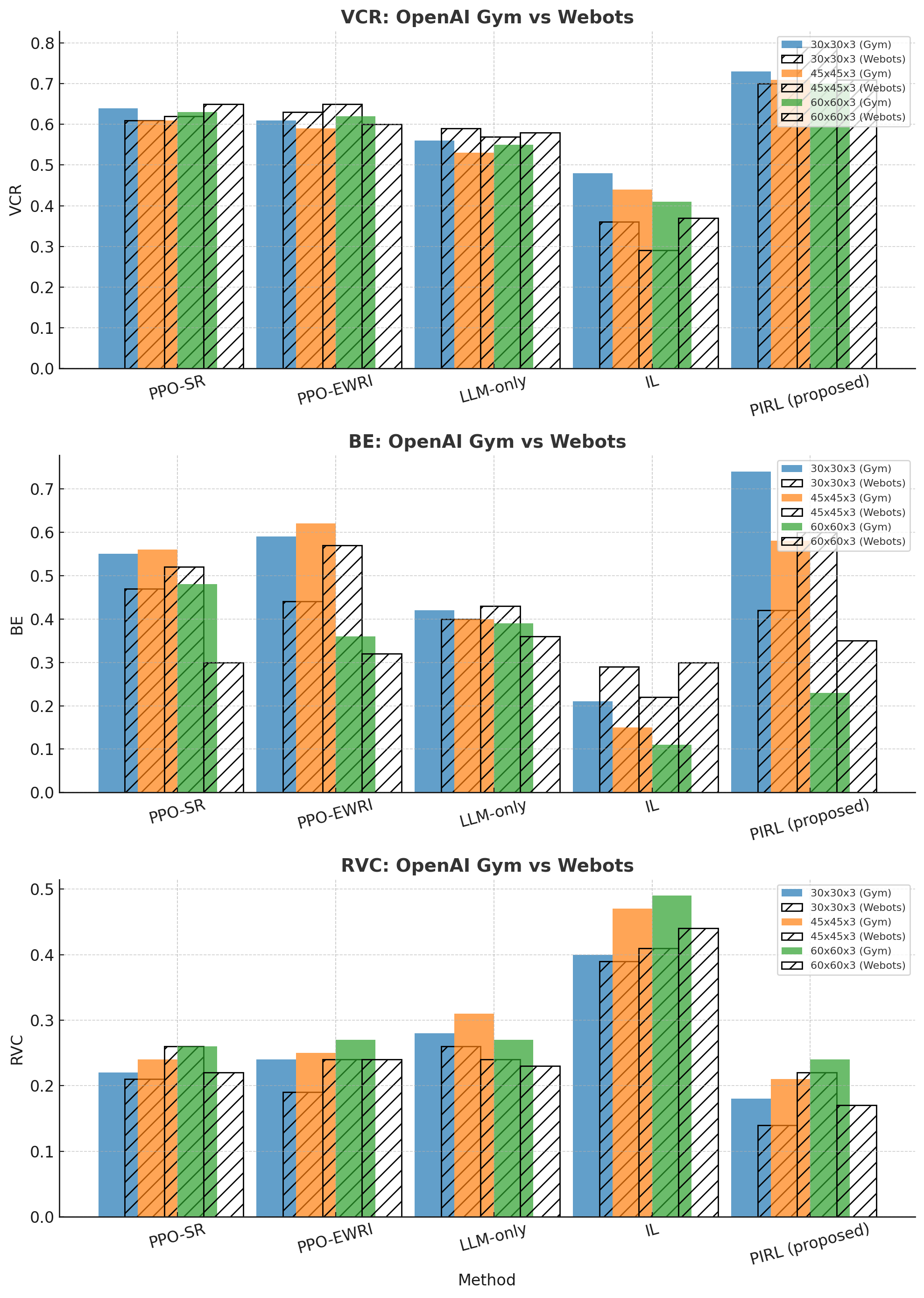}
    \caption{Results for VCR, BE, and RVC across Different Environments in OpenAI Gym and Webots}
    \label{fig:results}
\end{figure} 

\subsection{Limitations}
Although the proposed PIRL approach demonstrates robust performance across diverse environments, the approach comes with a few limitations. They are: 
\begin{itemize}
    \item PIRL relies on access to a pretrained LLM with zero-shot reasoning and in-context learning capabilities to generate semantic guidance during training. This dependency may introduce computational overhead and latency, which could be limiting for resource-constrained settings or cost-sensitive applications.
    \item The current formulation models the state space discretely for both position and camera parameters, which simplifies the learning problem but may limit the granularity of control and realism. Extending PIRL to continuous or hybrid state-action spaces would enable more precise control and facilitate real-world deployment. 
    \item While sim-to-real-like evaluation using Webots introduces physical realism, real-world deployment would require further considerations such as sensor noise, localization error, and computational constraints onboard the UAV platform.
\end{itemize}